\documentclass[11pt]{article}
\usepackage[preprint]{acl}
\usepackage{times}
\usepackage{latexsym}
\usepackage{todonotes}
\usepackage[T1]{fontenc}
\usepackage[utf8]{inputenc}
\usepackage{microtype}
\usepackage{inconsolata}
\usepackage{graphicx}
\usepackage[export]{adjustbox}
\usepackage{tabularx}
\usepackage{booktabs}
\usepackage{enumitem}
\usepackage{multirow}
\usepackage{multicol}
\usepackage{makecell}
\usepackage[table]{xcolor}
\usepackage{subcaption}
\newcommand{\perc}{\footnotesize{\%}}

\title{V-DyKnow: A Dynamic Benchmark for \\ Time-Sensitive Knowledge in Vision Language Models}

\author{\textbf{Seyed Mahed Mousavi\thanks{Equal contribution.}, Christian Moiola\textsuperscript{*}, Massimo Rizzoli\textsuperscript{*}, Simone Alghisi\textsuperscript{*},}\\ \textbf{Giuseppe Riccardi }\\
      Signals and Interactive Systems Lab, University of Trento, Italy\\
    \texttt{ \{mahed.mousavi, giuseppe.riccardi\}@unitn.it}
}

\begin{document}
\maketitle
\begin{abstract}
Vision-Language Models (VLMs) are trained on data snapshots of documents, including images and texts. Their training data and evaluation benchmarks are typically static, implicitly treating factual knowledge as time-invariant. However, real-world facts are intrinsically time-sensitive and subject to erratic and periodic changes, causing model predictions to become outdated. We present V-DyKnow, a Visual Dynamic Knowledge benchmark for evaluating time-sensitive factual knowledge in VLMs. Using V-DyKnow, we benchmark closed- and open-source VLMs and analyze \textit{a)} the reliability (correctness and consistency) of model responses across modalities and input perturbations; \textit{b)} the efficacy of knowledge editing and multi-modal RAG methods for knowledge updates across modalities; and \textit{c)} the sources of outdated predictions, through data and mechanistic analysis. Our results show that VLMs frequently output outdated facts, reflecting outdated snapshots used in the (pre-)training phase. Factual reliability degrades from textual to visual stimuli, even when entities are correctly recognized. Besides, existing alignment approaches fail to consistently update the models’ knowledge across modalities. Together, these findings highlight fundamental limitations in how current VLMs acquire and update time-sensitive knowledge across modalities. We release the benchmark, code, and evaluation data\footnote{Benchmark material available at \url{https://github.com/sislab-unitn/V-DyKnow}}.
\end{abstract}

\section{Introduction}

\begin{figure}[t]
    \centering
        \includegraphics[width=\linewidth]{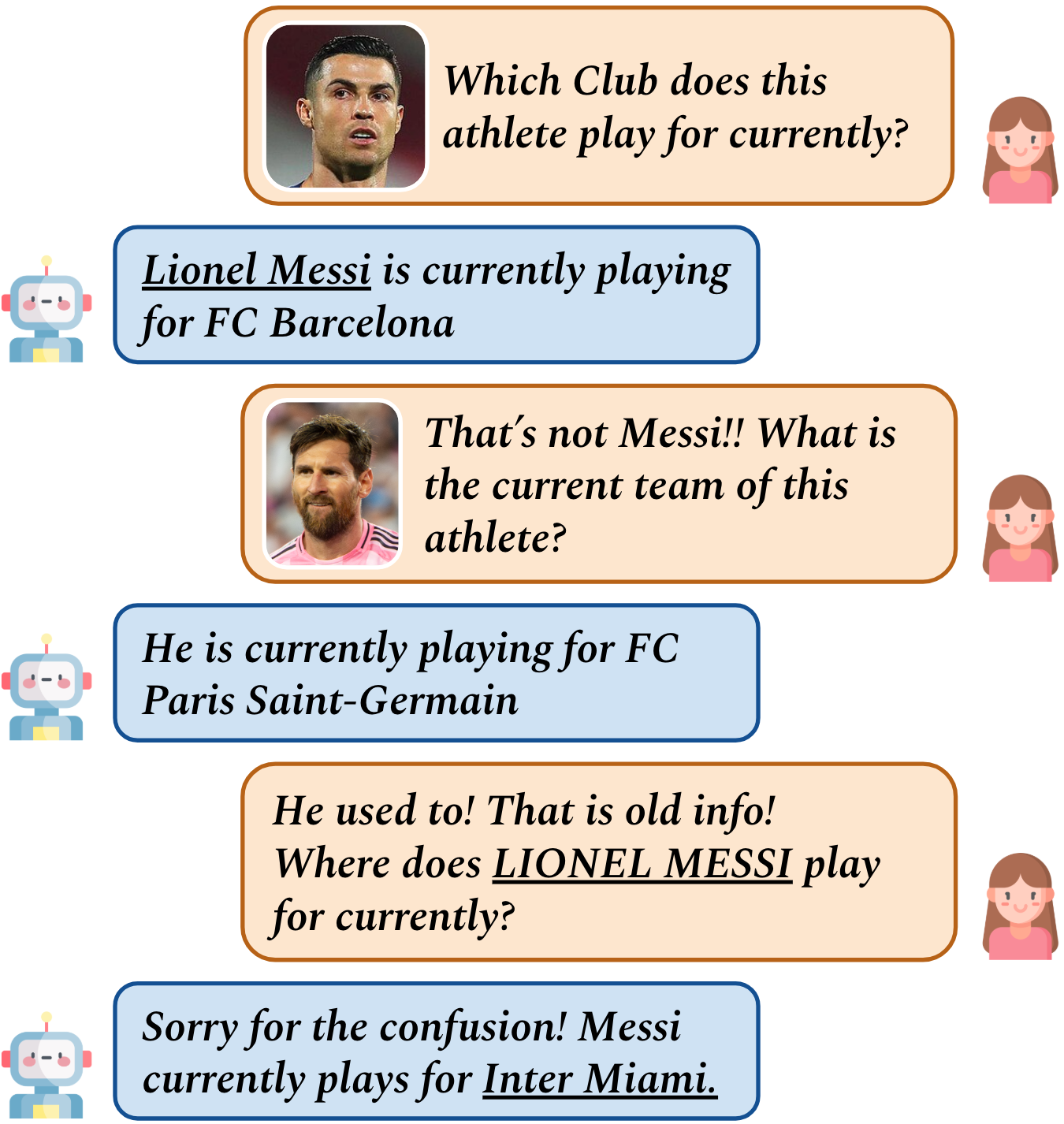} 
    \caption{An example of multimodal querying VLMs for factual knowledge that is time-sensitive. Upon a visual stimulus, the VLM first misidentifies the entity and retrieves an incorrect fact. Following the clarification turn about the entity, the model generates an outdated answer. In the final turn, the entity is explicitly stated in the text, and the correct fact is finally returned. This example highlights key issues investigated in this work: the prevalence of outdated knowledge in VLMs and the performance gap in VLMs for visual and textual stimuli.}
    \label{fig:firstfig}
\end{figure}

Vision-Language Models (VLM) are increasingly used for answering factual questions about the real world from visual inputs. In these settings, VLMs are expected to connect the images' content with stored world knowledge and output answers to the question~\cite{cohen-etal-2025-performance}. Similar to Large Language Models (LLMs), VLMs are static models whose factual knowledge is acquired during training from large heterogeneous data snapshots collected at different points in time. These snapshots capture different states of the world and its entities and may encode outdated and/or conflicting versions of the same facts. As a result, the factual knowledge expressed by a VLM reflects the content and temporal mixture of its (pre-)training data, rather than the state of the world at inference time.

The evaluation of knowledge in VLMs is predominantly conducted using static benchmarks with fixed ground truth \cite{cohen-etal-2025-performance, papi2026mcif}. Prior work has examined factual knowledge in VLMs and identified modality-dependent gaps between visual and textual entity representations~\cite{cohen-etal-2025-performance}. Research on multimodal knowledge editing has studied whether factual updates transfer across modalities \cite{Fang_2025_ICCV, Chen_2025_CVPR}. However, these efforts focus on a fixed set of static facts, implicitly assuming that factual correctness is time-invariant. As real-world facts evolve, such benchmarks fall out of sync with the current state of the world, and the evaluation scores thus measure agreement with outdated information rather than factual correctness at inference time, undermining the validity of the evaluation itself. This misalignment is particularly severe in visual settings, where model responses depend jointly on entity recognition, cross-modal understanding, and factual recall, similiar to the example in Figure \ref{fig:firstfig}.

To enable the evaluation of time-sensitive factual knowledge in VLMs, we introduce V-DyKnow, a visual dynamic benchmark for VLMs. Instead of relying on fixed annotations, model outputs are validated against up-to-date factual information derived from Wikidata \cite{10.1145/2629489} at evaluation time, allowing correctness to be assessed relative to the current state of the world. V-DyKnow represents factual queries as visual entities (e.g., portraits, flags, and logos) paired with relations and attributes, each associated with temporal validity intervals. V-DyKnow supports analysis of outdatedness, cross-modality evaluation, and input-bound and cross-modal consistency.

Using V-DyKnow, we conduct a systematic analysis of time-sensitive factual knowledge in VLMs along three axes. First, we evaluate how accurately and consistently VLMs retrieve real-world facts when entities are presented visually rather than textually, and quantify the prevalence of outdated responses under visual grounding. Second, we examine whether existing alignment approaches, including knowledge editing and multimodal retrieval-augmented generation (RAG), can update outdated facts across modalities. And third, we investigate the sources of outdated predictions through training-data analysis and mechanistic interpretability. Our results show that VLMs frequently retrieve outdated facts and that factual recall degrades under visual stimulus, even when the entity is correctly recognized. Existing updating methods work only in simplified scenarios and interfere with knowledge acquired during pre-training. Together, these findings highlight limitations in how current VLMs acquire, recall, and maintain time-sensitive knowledge. Our contributions can be summarized as follows:

\begin{itemize} [noitemsep,topsep=1pt,parsep=1pt,partopsep=1pt]
    \item We introduce \textbf{V-DyKnow}, a dynamic benchmark for evaluating time-sensitive factual knowledge in Vision–Language Models across modality and input perturbations.
    \item We provide the first systematic evaluation of time-sensitive factual knowledge in state-of-the-art VLMs, showing the prevalence of outdated facts and large performance gap across modalities.
    \item We analyze the effectiveness of existing alignment approaches, including knowledge editing and multimodal retrieval-augmented generation.
    \item We investigate the sources of outdated predictions through data and mechanistic analyses, linking model outputs to training data snapshots and examining how factual recall emerges across model layers.
\end{itemize}

\section{V-DyKnow}
\label{sec:v-dyknow}

Building on DyKnow \cite{mousavi-etal-2024-dyknow}, we introduce V-DyKnow, a benchmark for evaluating time-sensitive factual knowledge in VLMs under both textual and visual grounding.

\begin{table*}[t]
    \centering
    \small
    \begin{adjustbox}{width=\linewidth,center=\linewidth}
        \begin{tabular}{lrrrccrrrccrrr}
            
            \toprule
            \multirow{2}{*}{\textbf{(Year) Model}} & \multicolumn{3}{c}{\textbf{VLM Visual Prompt}} & & &
            \multicolumn{3}{c}{\textbf{VLM Textual Prompt}} & & & \multicolumn{3}{c}{\textbf{LLM* Textual Prompt}} \\
            \cmidrule(l{4pt}r{4pt}){2-4} \cmidrule(l{4pt}r{4pt}){7-9} \cmidrule(l{4pt}r{4pt}){12-14}
            & \textbf{C}{\scriptsize orrect}& \textbf{O}{\scriptsize utdated} &\textbf{I}{\scriptsize rrelevant} & & &
            \textbf{C}{\scriptsize orrect}& \textbf{O}{\scriptsize utdated} &\textbf{I}{\scriptsize rrelevant} & & &
            \textbf{C}{\scriptsize orrect}& \textbf{O}{\scriptsize utdated} &\textbf{I}{\scriptsize rrelevant} \\
            
            \midrule
            \rowcolor{gray!15}
            {\small(2023) LLaVA-1.5}       & 13\% & 31\% & 56\% & & & 33\% & 49\% & 18\%  & & & 33\% & 47\% & 19\%\\
            {\small(2024) LLaVA-OneVision} & 22\% & 36\% & 42\% & & & 31\% & 45\% & 24\%  & & & 37\% & 42\% & 22\% \\
            \rowcolor{gray!15}
            {\small(2024) PaliGemma 2}     & 3\%  & 4\%  & 93\% & & & 0\%  &  0\% & 100\% & & & 62\% & 28\% & 10\% \\
            {\small(2024) Molmo}           & 24\% & 20\% & 56\% & & & 34\% & 44\% & 22\%  & & & 40\% & 40\% & 20\%\\
            \rowcolor{gray!15}
            {\small(2024) Qwen2-VL}        & 28\% & 38\% & 34\% & & & 36\% & 44\% & 20\%  & & & 37\% & 42\% & 22\% \\
            {\small(2025) Qwen2.5-VL}      & 32\% & 38\% & 30\% & & & 39\% & 40\% & 22\%  & & & 44\% & 35\% & 21\%\\
            \rowcolor{gray!15}
            {\small(2025) InternVL3.5}     & 26\% & 21\% & 53\% & & & 40\% & 29\% & 31\%  & & & 51\% & 29\% & 19\%\\
            {\small(2025) GPT-4}           & 71\% & 18\% & 11\% & & & 72\% & 19\% & 9\%   & & & -&-&- \\
            \rowcolor{gray!15}
            {\small(2025) GPT-5}           & 75\% & 15\% & 10\% & & & 76\% & 14\% & 9\%   & & & -&-&-\\
            
            \bottomrule
        \end{tabular}
    \end{adjustbox}
    \caption{Evaluation of state-of-the-art VLMs on V-DyKnow under visual and textual prompting. Results report the percentage of \textbf{C}orrect (currently valid attribute), \textbf{O}utdated (historically valid but no longer current), and \textbf{I}rrelevant (neither current nor historically valid) responses using the \textit{upper-bound} strategy. The results can be compared with the performance of the corresponding LLM prior to multimodal (pre-)training/instruction tuning (\textbf{LLM* Textual Prompt}).}
    \label{table:visual-vs-text}
\end{table*}

\textbf{Facts} Following the prior work \cite{mousavi-etal-2024-dyknow,mousavi2025llms}, we represent factual knowledge as \textit{(subject, property, attribute)} triplets. The \textit{subject} denotes a real-world entity (e.g., \textit{Apple}), the \textit{property} defines a relation associated with that entity (e.g., \textit{CEO}), and the \textit{attribute} corresponds to the value of that relation (e.g., \textit{Tim Cook}). To construct V-DyKnow, we extract the up-to-date attribute value for each subject-property in DyKnow from Wikidata \cite{10.1145/2629489}, a continuously updated knowledge graph. Attributes in Wikidata are connected to subjects through properties and enriched with qualifiers specifying their validity intervals. For example, the subject \textit{Apple} is associated with attribute \textit{Tim Cook} via the property \textit{CEO}, annotated with the validity interval \textit{2011–present}. Meanwhile, outdated attributes are preserved with their previously valid time spans (e.g., \textit{Steve Jobs 1997-2011}). The final benchmark contains 139 time-sensitive facts, including 82 facts associated with 47 countries, 28 facts related to athletes, and 29 facts corresponding to 22 organizations (i.e., companies and institutions).

\textbf{Visual Prompts} To evaluate factual knowledge across modalities, we construct paired queries in which the same subject entity is presented either textually or visually. In the textual setting, the subject appears explicitly in the prompt (e.g., \textit{Who is the CEO of Apple?}). In the visual setting, the textual subject is replaced by the corresponding image of the entity obtained from Wikidata (e.g., \textit{Who is the CEO of this company?}). This formulation enables a direct comparison of factual recall when entities are accessed through linguistic versus visual representation. The visual representations consist of coats of arms and flags for countries, portraits for athletes, and logos for organizations. For countries, we include the coat of arms to help mitigate potential ambiguity, as certain flags share similar color patterns and may become difficult to distinguish under transformations\footnote{For example, the flag of the Netherlands, rotated by 90° resembles the French flag. The coat of arms, therefore, provides an additional visual cue that uniquely identifies the country.}. The prompts are presented in § Table~\ref{tab:visual_prompts}.

\textbf{Visual Entity Recognition} Visual queries introduce an additional challenge compared to text-only prompts, since the model must first identify the depicted entity and then retrieve the corresponding attribute from its internal knowledge. When a model produces an incorrect answer, it is therefore unclear whether the error arises from a failure of visual recognition or from missing factual knowledge. To disentangle these factors, we introduce an auxiliary recognition task, where models are prompted to identify the depicted entity. Prompt templates for the detection task are generated using the same procedure as the factual queries, presented in § Table \ref{tab:detection_prompts}. 

\textbf{Protocol} For each query, the model’s response is compared against the attribute collected at the evaluation time, and classified as \textbf{\textit{Correct}} if it matches the currently valid attribute. It is labeled \textbf{\textit{Outdated}} if it corresponds to an attribute that was valid in the past. Finally, a response is categorized as \textbf{\textit{Irrelevant}} if it matches neither the current nor any previously valid attribute associated with the queried property. This fine-grained distinction of the errors enables a more precise characterization of how VLMs encode and retrieve time-sensitive information.

\textbf{Strategy} Generative models are known to be sensitive to prompt lexicalization, where different versions of the same query may lead to different responses despite identical semantics. To address this limitation, V-DyKnow evaluates each factual query using three prompts with minor perturbations in lexicalization while preserving the same semantics (presented in § Table~\ref{tab:visual_prompts}). Each model is queried independently with the three prompts, and the responses are evaluated under the \textit{Upper-Bound} strategy, where the final prediction for a query corresponds to the best-performing response among the three prompts (Correct > Outdated > Irrelevant). This procedure reduces the likelihood that evaluation outcomes are driven by prompt-specific artifacts rather than by the model’s underlying factual knowledge.

\textbf{Models} We evaluate 9 VLMs (and their corresponding LLM): LLaVA-1.5 7B (Vicuna-v1.5 7B) \cite{Liu_2024_CVPR} LLaVA-OneVision 7B (Qwen2 7B) \cite{li2025llavaonevision}, PaliGemma 2 10B (Gemma 2 9B) \cite{steiner2024paligemma}, Molmo 7B (OLMo 7B) \cite{Deitke_2025_CVPR}, Qwen2-VL 7B (Qwen2 7B) \cite{wang2024qwen2}, Qwen2.5-VL 7B (Qwen2.5 7B) \cite{qwen2025qwen25technicalreport}, InternVL3.5 8B (Qwen3 8B) \cite{wang2025internvl3}, GPT-4\footnote{\href{https://developers.openai.com/api/docs/models/gpt-4.1}{gpt-4.1-2025-04-14}}, and GPT-5\footnote{\href{https://developers.openai.com/api/docs/models/gpt-5.1}{gpt-5.1-2025-11-13}}. Additional details about the model's checkpoints are presented in § Table \ref{tab:model_checkpoints}

\begin{table}[t!]
\centering
\small
\begin{tabular}{lr}
    \toprule
    \textbf{(Year) Model} & \textbf{Accuracy} \\
    \midrule
    {\small(2023)} LLaVA-1.5 & 69\% \\
    {\small(2024)} LLaVA-OneVision  & 80\% \\
    {\small(2024)} PaliGemma 2  & 63\% \\
    {\small(2024)} Molmo & 60\% \\
    {\small(2024)} Qwen2-VL & 91\% \\
    {\small(2025)} Qwen2.5-VL & 87\% \\
    {\small(2025)} InternVL3.5 & 58\% \\
    {\small(2025)} GPT-4 & 83\% \\
    {\small(2025)} GPT-5 &  75\% \\
    \bottomrule
\end{tabular}
\caption{Accuracy on the Visual Entity Recognition task. Models are prompted to identify the entity depicted in each image before answering the factual query. Scores are computed using an \textit{Upper-Bound} across three prompt variants (presented in § Table~\ref{tab:visual_prompts}).}
\label{table:upper-bound}
\end{table}

\section{Results}
We assess the time-sensitive knowledge in state-of-the-art VLMs on V-DyKnow. We analyze how frequently models produce outdated responses, their performance under minor perturbations, and the effectiveness of existing methods for updating their factual knowledge.

\subsection*{A. Model Evaluation}
Table~\ref{table:visual-vs-text} reports the performance of nine state-of-the-art VLMs on V-DyKnow\footnote{The models are assessed against the factual knowledge retrieved on Nov 2025 from Wikidata.}. Across models, outdated factual responses appear frequently, often exceeding the number of correct answers. Many models retrieve the knowledge that no longer reflects the current state of the world. For most models, factual retrieval is substantially more reliable when the entity is provided textually. This gap is more evident in LLaVA-1.5, Molmo, InternVL3.5, and the Qwen family, resulting in a large proportion of irrelevant responses upon visual stimuli. This gap suggests that factual knowledge accessible through textual queries is not always accessible when the same entity must first be recognized from an image, highlighting a limitation in the interaction between visual recognition and factual recall. While more recent models (e.g., Qwen2.5-VL, GPT-\{4,5\}) achieve higher rates of correct answers, outdated responses continue to constitute a substantial fraction of model outputs. Notably, the gap between visual and textual prompting is substantially reduced in the most recent models. Comparing VLMs with their corresponding base LLM reveals that multimodal training often degrades factual recall. Except for LLaVA-1.5, the textual performance of the VLM is consistently lower than its corresponding LLM, suggesting that multimodal alignment can obscure knowledge that is otherwise accessible in the underlying language model. PaliGemma 2 represents an extreme case: after multimodal alignment, the model outputs ``unanswerable'' to most of our queries.

\textbf{Visual Entity Recognition} The results, reported in Table~\ref{table:upper-bound}, indicate that models with a smaller gap between visual and textual factual retrieval exhibit a strong recognition performance. The Qwen2 model family achieves entity recognition accuracy above 85\%, and correspondingly shows a moderate difference between visual and textual prompting in Table~\ref{table:visual-vs-text}. In contrast, models with weaker entity recognition performance such as LLaVA-1.5, Molmo, and InternVL3.5 tend to display substantially larger modality gaps, with higher proportions of irrelevant outputs when the entity is presented visually. These results suggest that failures in visual entity recognition contribute to the degradation of factual retrieval under visual prompting. Nevertheless, even models with strong recognition capabilities, such as Qwen2-VL and Qwen2.5-VL, still produce a large fraction of outdated answers, indicating that correctly identifying the entity does not guarantee access to up-to-date factual knowledge.

\begin{table}[t!]
    \centering
    \small
    \begin{tabular}{lrr}
        \toprule
        \multirow{2}{*}{\textbf{(Year) Model}} & \multicolumn{2}{c}{\textbf{Prompt Agreement}} \\
        \cmidrule(lr){2-3}
        & \textit{Visual} & \textit{Textual} \\
        \midrule
        {\small(2023)} LLaVA-1.5       & 52\% & 70\% \\
        {\small(2024)} LLaVA-OneVision & 66\% & 80\% \\
        {\small(2024)} PaliGemma 2     & 25\% & 100\% \\
        {\small(2024)} Molmo           & 57\% & 81\% \\
        {\small(2024)} Qwen2-VL        & 46\% & 79\% \\
        {\small(2025)} Qwen2.5-VL      & 70\% & 78\% \\
        {\small(2025)} InternVL3.5     & 58\% & 75\% \\
        {\small(2025)} GPT-4           & 92\% & 97\% \\
        {\small(2025)} GPT-5           & 84\% & 90\% \\
        \bottomrule
    \end{tabular}
    \caption{Output consistency (prompt agreement) across three different lexicalizations of the same query for \textit{Visual} and \textit{Textual} inputs. An agreement measures the percentage of cases for which all prompts produce the same answer (i.e., correct, outdated, or irrelevant).}
    \label{table:prompt-agreement}
\end{table}

\begin{table*}[t!]
\centering
\small
\setlength{\tabcolsep}{6pt}
\renewcommand{\arraystretch}{1.2}
\begin{tabular}{l  c  c c c  c c}
\toprule
\multirow{3}{*}{\textbf{(Year) Model}} & \multirow{3}{*}{\shortstack{\textbf{\# Outdated} \\ \textbf{Facts}}} & \multicolumn{3}{c}{\textbf{Knowledge Editing}} & \multicolumn{2}{c}{\textbf{Multi-modal RAG}} \\ \cmidrule(lr){3-5} \cmidrule(lr){6-7}
 & & \textbf{WISE} & \textbf{GRACE} & \textbf{IKE} & \textbf{Retrieved Doc.} & \textbf{Gold Doc.} \\
\midrule
(2023) LLaVA-1.5 & 69 & 2.9\% & 5.5\% & 95.6\% & 73.5\% & 84.4\% \\
(2024) Qwen2-VL & 83 & 3.9\% & 2.4\% & 100.0\% & 80.1\% & 92.8\% \\
\bottomrule
\end{tabular}
\caption{Effectiveness of existing alignment methods (editing and multimodal RAG) on outdated facts for LLaVA-1.5 and Qwen2-VL. Scores report the harmonic mean of efficacy and paraphrase success. We compare knowledge editing methods with multimodal RAG using either retrieved documents as well as gold documents.}
\label{tab:knowledge_editing_rag_vlm}
\end{table*}

\subsection*{B. Output Consistency} 
We evaluate the consistency of model outputs across different lexicalizations of the same query as prompt agreement \cite{mousavi2025llms}. Across models, presented in Table~\ref{table:prompt-agreement}, agreement is consistently higher for textual prompts, indicating that predictions are more sensitive to prompt formulation when the entity must first be inferred from an image. This pattern is particularly pronounced for models such as LLaVA-1.5, Molmo, and Qwen2-VL. In contrast, models such as GPT-4 and GPT-5 exhibit high agreement across both modalities, indicating more stable retrieval behavior. An extreme case is PaliGemma, which consistently generates irrelevant answers. Overall, these results indicate that factual retrieval in VLMs is not only modality-dependent but also sensitive to prompt formalization.

\subsection*{C. Updating VLMs' Knowledge}

The precedence of outdated factual knowledge in VLMs,Table~\ref{table:visual-vs-text}, motivates evaluating whether existing alignment approaches can effectively update the factual knowledge encoded in VLMs. We evaluate knowledge editing methods, aiming to update factual knowledge without retraining the full model. We consider three representative approaches: \textbf{GRACE}~\cite{hartvigsen2023aging}, which stores edits in an external memory module applied through an adaptor; \textbf{WISE}~\cite{wang2024wise}, which extends GRACE by also storing pre-trained knowledge in an external memory module; and \textbf{IKE}~\cite{zheng2023can}, which introduces updated examples through in-context learning. We compare the editing methods with multimodal Retrieval-Augmented Generation (RAG) \cite{das2025training} under two settings: a) a realistic setting where documents are retrieved based on the input question; and b) an oracle setting where the gold document containing the answer is provided directly to the model. We focus on LLaVA-1.5 and Qwen2-VL, as the old models with the highest percentage of outdated responses included in Table~\ref{table:visual-vs-text}. We measure the effectiveness of each approach using the harmonic mean of efficacy and paraphrase success \cite{mousavi-etal-2024-dyknow, mousavi2025llms}. Additional implementation details are presented in Sections §\ref{app:editing_methods}, §\ref{mrag}, and §\ref{subsec:metrics} 

Table~\ref{tab:knowledge_editing_rag_vlm} shows that performance varies substantially across methods.  Among the knowledge editing methods, only IKE achieves a high harmonic mean, whereas WISE and GRACE perform poorly (both below 6\%). However, IKE is not fully realistic, as it requires the deterministic up-to-date fact for each query, whereas facts should be retrieved based on the question (as done in multimodal RAG). Retrieval-based approaches show more consistent performance. Multimodal RAG achieves high scores across both models, although performance depends on retrieval quality, as reflected by the gap between retrieved and gold documents. Importantly, both RAG and IKE rely on external information at inference time and therefore do not update the model’s parametric knowledge. As a result, both methods may still suffer from inconsistencies between the model’s internal knowledge and the external information provided at inference time~\cite{wu2024clasheval}.

\begin{table}[t!]
    \centering
    \small
    \begin{tabular}{lrrrr}
        \toprule
        \multirow{2}{*}{\textbf{Model}} & \multirow{2}{*}{\textbf{C}{\scriptsize orrect}} & \multicolumn{3}{c}{\textbf{Incorrect}}\\
        \cmidrule(lr){3-5}
        & & \textbf{O}{\scriptsize utdated} & \textbf{G}{\scriptsize eneric} & \textbf{H}{\scriptsize alluc.}\\
        \midrule
        {\textbf{LLaVA-1.5}} & & & &  \\
          \hspace{0.1cm} WISE & 3\% & 73\% & 0\% & 24\% \\
          \hspace{0.1cm} GRACE & 21\% & 63\% & 1\% & 15\%  \\
          \hspace{0.1cm} IKE & 95\% & 0\% & 5\% & 0\%  \\
          \hspace{0.1cm} RAG & & & &  \\
            \hspace{0.3cm} {\textit{Ret. Doc}} & 74\% & 2\% & 0\% & 23\%  \\
            \hspace{0.3cm} {\textit{Gold Doc}} & 85\% & 3\% & 0\% & 12\%  \\
        \midrule
        {\textbf{Qwen2-VL}} & & & &  \\
          \hspace{0.1cm} WISE & 4\% & 34\% & 3\% & 57\%  \\
          \hspace{0.1cm} GRACE & 34\% & 51\% & 0\% & 14\% \\
          \hspace{0.1cm} IKE & 100\% & 0\% & 0\% & 0\%  \\
          \hspace{0.1cm} RAG & & & &  \\
          \hspace{0.3cm} {\textit{Ret. Doc}} & 80\% & 5\% & 0\% & 15\% \\
          \hspace{0.3cm} {\textit{Gold Doc}} & 93\% & 3\% & 0\% & 4\%  \\
        
        \bottomrule
    \end{tabular}
\caption{Qualitative error analysis of knowledge editing (WISE, GRACE, IKE) and retrieval-based alignment methods across two representative LLMs, based on human annotation of generated outputs. Responses are categorized as \textbf{Correct} when the outdated attribute is successfully updated. Incorrect responses are further classified as \textbf{Outdated} when the model continues to return the outdated fact, \textbf{Generic} when the model produces vague answers, and \textbf{Hallucinated} when the model fabricates or confuses attributes.}
\label{tab:editing_error_breakdown}
\end{table}

\textbf{Qualitative Analysis} To better understand the impact of alignment approaches on the model outputs, we annotate the generated responses using a human evaluation protocol adapted from \citet{mousavi2022evaluation}. Two researchers independently reviewed the outputs produced after editing/RAG. Responses are annotated as \textbf{C}{\scriptsize orrect} if a previously outdated answer is updated successfully. Meanwhile, \textbf{I}{\scriptsize ncorrect} outputs are further subdivided into a) \textbf{O}{\scriptsize utdated}, when the model continues to return outdated knowledge; b) \textbf{G}{\scriptsize eneric}, if the model produces vague statements (e.g., ``The CEO of Apple is a very important person''); and c) \textbf{H}{\scriptsize allucination}, when the model fabricates or confuses attributes, such as attributing political roles to athletes. 

The results in Table~\ref{tab:editing_error_breakdown} further confirm that WISE and GRACE fail to reliably update the models’ factual knowledge, as the majority of their responses remain outdated even after editing. Both methods increase the proportion of hallucinated responses, with models either assigning incorrect attributes to a given subject-property pair (e.g., the emperor of Japan is Barack Obama) or answering with the name of the visual subject (e.g., \textit{``Can you name the current prime minister of this country? \underline{Italy}''}). IKE produces few generic outputs, such as answering \textit{``yes''} to questions like \textit{``can you name...''}. Finally, even when provided with gold documents, RAG can still produce outdated or hallucinated answers, confirming that the model’s pre-trained parametric knowledge can interfere with external evidence.

\section{Further Analysis}
We conduct additional analyses to better understand the sources of outdated responses. First, we examine the temporal validity intervals associated with model predictions to approximate the time period reflected in their parametric knowledge. We then analyze the training data of a representative model to determine whether the information required to answer a query was present in the pretraining corpus. Finally, we perform a mechanistic analysis to study how factual information is retrieved across model layers. 

\begin{figure*}[t]
    \centering
    \includegraphics[width=\linewidth]{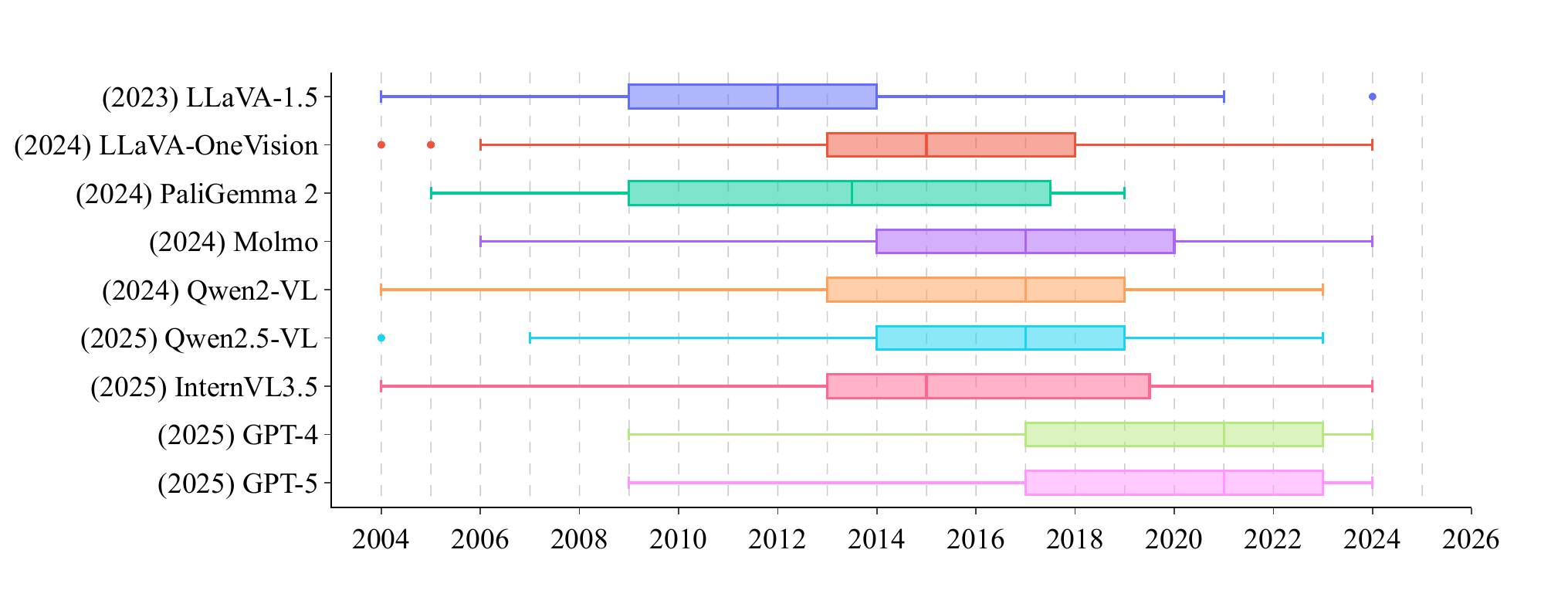}
    \caption{Temporal distribution of the model responses based on Wikidata. For each VLM, we map their correct and outdated responses (e.g., ``The CEO of Apple is \texttt{Steve Jobs}'') to the time at which the corresponding attribute was valid (e.g., ``1997-2011''). By aggregating these intervals using a boxplot, we can approximate the state of the world encoded in the model's parameters. For example, results show that, while Qwen2-VL responses range between 2004 and 2023 (in line with its reported cutoff), most of them are concentrated between 2013 and 2019.}
    \label{fig:data_interval_approximation}
\end{figure*}

\begin{figure*}[t]
    \centering
    \includegraphics[width=\linewidth]{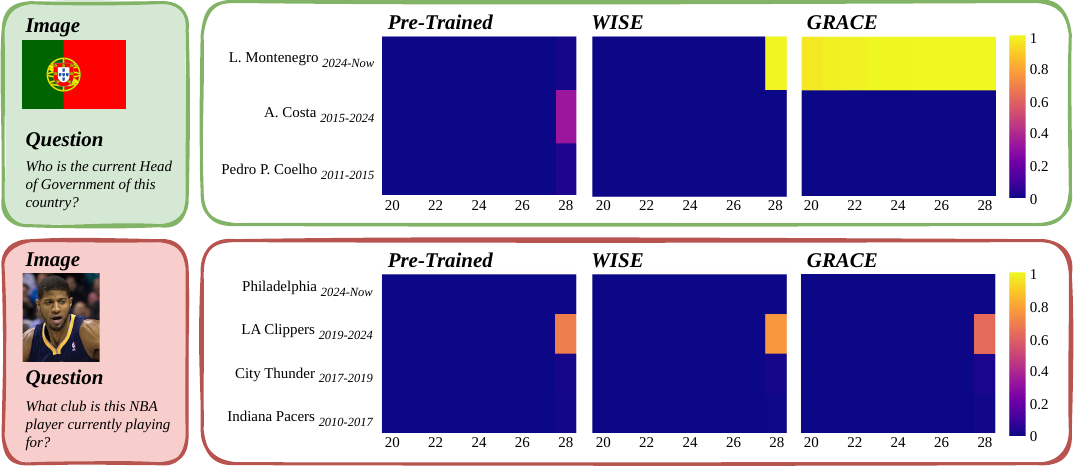}
    \caption{Mechanistic analysis of Qwen2-VL illustrating how editing methods modify the probability for certain attributes (y-axis) across different layers (x-axis) when asked about an Image-Question pair. In case of a successful update (top), WISE and GRACE affect different layers: WISE primarily edits the final layer, whereas GRACE modifies a broader range. In contrast, neither method successfully updates the basketball team of Paul George (bottom), suggesting that this knowledge may be more deeply embedded or that other stored facts may interfere.}
    \label{fig:mechanistic}
\end{figure*}

\subsection*{A. Data Interval Approximation}

Using the temporal validity intervals associated with Wikidata attributes, we analyze the correct and outdated outputs of each model to approximate the temporal interval reflected in their parametric knowledge. For each predicted attribute, we retrieve its validity interval and aggregate the distribution of these intervals across model outputs. The results are summarized in Figure~\ref{fig:data_interval_approximation}. Our approximations are in line with the cutoffs of the models with disclosed data intervals such as Qwen2-VL (June 2023)\footnote{\href{https://qwen-ai.chat/models/qwen2-vl/}{Qwen AI Chat Models}}, GPT-4 (June 1st 2024) and GPT-5 (September 30th 2024) API\footnote{\href{https://developers.openai.com/api/docs/models}{OpenAI API Models}}.
Although all evaluated models were released after 2023, most responses correspond to facts valid before 2020, suggesting that the underlying training data predominantly reflects earlier states of the world. Within several model families, newer versions tend to generate responses associated with more recent intervals. For example, LLaVA-OneVision and Qwen2.5-VL produce fewer responses tied to earlier years than LLaVA-1.5 and Qwen2-VL, respectively. Across open-source models, the temporal intervals roughly starts from 2005. GPT-\{4,5\} produce responses concentrated in more recent intervals, primarily between 2017 and 2023, although outdated predictions can still extend to earlier periods. Notably, both GPT models show nearly identical temporal distributions, suggesting that they rely on similar training data snapshots. Overall, these patterns indicate that most of the model outputs and presumably thier training data reflect an outdated state of the world, even though more recent models tend to incorporate more recent information.

\subsection*{B. Linking Responses to Training Data }
To understand how VLMs' predictions are shaped by the information available during pretraining, we analyze the training corpus for a stratified subset of entities. We conduct this analysis using Molmo, as its pretraining data is publicly available and therefore allows direct inspection of the documents the model has been exposed to. We focus on the Wikipedia portion of the corpus, as prior work suggests that models tend to acquire factual knowledge more reliably from structured and curated sources such as Wikipedia compared to other documents in the pre-training snapshots\cite{li2024formality}. For each entity, we analyze the main Wikipedia page included in the pretraining snapshot and determine whether the current fact appears in the document and how prominently it is mentioned. Our goal is not to exhaustively search all documents in the corpus, but to determine whether the correct information was present in the canonical source at the time the training snapshot was collected. If the correct information appears in this version of the page, it is likely that the model was exposed to it during pretraining; conversely, if it is absent, the model may have only observed the outdated attribute or no relevant information at all. 

We select 30 entities, 10 cases where Molmo generates the correct attributes, 10 where it produces an outdated attribute, and 10 where it generates an irrelevant response. For entities where Molmo produces the correct answer, in five cases the correct fact does not appear in the document, suggesting that the model must have acquired this information from other sources in its pretraining data. In one case, the correct attribute is mentioned but not as the most frequent value, while in four cases, it is both present and the most frequently occurring information in the page. In cases where Molmo produces outdated answers, in seven cases the current information does not appear in the Wikipedia page, indicating that the model’s prediction is consistent with the information available in the document at the time of pretraining. However, in three cases, the current attribute is present and even the most frequently occurring value in the page, yet the model still retrieves the outdated one. This suggests that outdated predictions cannot always be explained by missing information in the training data. Finally, for entities where Molmo produces irrelevant responses, the current attribute is absent from the Wikipedia page in six cases. In the remaining four cases, the correct face appears in the document, twice as a secondary mention and twice as the most frequent value, yet the model still fails to retrieve it. Taken together, these observations indicate that model behavior is not determined solely by the presence or frequency of information in the training data.

\subsection*{C. Mechanistic Interpretability}
\label{subsec:mechanistic_interpretability}
We conduct a mechanistic analysis to examine how different editing methods modify the retrieval of factual information across model layers. Building on prior work~\cite{mousavi2025llms, geva-etal-2023-dissecting, ou-etal-2025-llms}, we probe the model with visual queries and analyze how different layers contribute to the prediction of updated and outdated attributes. Figure~\ref{fig:mechanistic} presents the results of this analysis for Qwen2-VL. Additional figures and detailed analyses are provided in Section §\ref{subsec:app-mechanistic}. Overall, we observe that the pre-trained model assigns a high probability to one of the candidate attributes only at the final layer, suggesting that earlier layers contribute little to the explicit factual recall reflected in the final prediction. In case of successful edits (the upper part in Figure~\ref{fig:mechanistic}), while WISE only modifies the final layer, GRACE targets earlier layers and propagates the change across layers 20 to 28. Meanwhile, in the case of unsuccessful edits (the lower part in Figure~\ref{fig:mechanistic}), GRACE slightly reduces the probability assigned to the incorrect entity, whereas WISE increases it, potentially due to interference with other stored factual associations.

\section{Literature Review}

\textbf{Benchmarks} Existing evaluations of time-sensitive knowledge have been largely restricted to the textual inputs. Dynamic benchmarks such as DyKnow \cite{mousavi-etal-2024-dyknow}, EvolveBench \cite{zhu-etal-2025-evolvebench}, and DynaQuest \cite{lin-etal-2025-dynaquest} have been introduced to assess how models handle temporally changing facts, but these are designed exclusively for Large Language Models (LLMs). In contrast, existing benchmarks with visual queries \cite{chen-etal-2023-pre-trained, Cheng_2025_ICCV} are largely static, making it difficult to assess the validity of time-sensitive facts under visual grounding. 

\textbf{VLMs} Current research suggests that VLMs suffer from significant representational discrepancies, making them unreliable repositories of knowledge \cite{huh2024platonicrepresentationhypothesis}. Recent studies identify a huge performance gap across modalities, with high performance with textual queries \citet{cohen-etal-2025-performance}, indicating the VLMs fail to utilize the visual signals effectively \citet{fu2025hidden}. 
This inconsistency has also been observed in video and speech \cite{papi2026mcif}, and in multimodal knowledge editing \cite{Fang_2025_ICCV}, where edits fail to transfer across modalities. Meanwhile, multimodal knowledge editing studies rely on static ground truth and do not take in consideration the temporal validity of facts or factual updates over time \cite{Fang_2025_ICCV, Chen_2025_CVPR}.

\section{Discussion \& Conclusion}

Our analysis shows that outdated knowledge is common across both open- and closed-source VLMs, reflecting old training data snapshots even in recent models. These findings raise broader questions about how large multimodal models should represent knowledge about a world that continuously evolves. Current training paradigms rely on static data snapshots, while existing alignment methods remain insufficient for maintaining consistent and up-to-date factual knowledge. Addressing these challenges may require new learning paradigms that explicitly model temporal validity, integrate dynamic knowledge sources, and support continual updates without disrupting existing representations.
We believe V-DyKnow provides a useful resource for studying these questions. By explicitly modeling temporal validity and supporting multimodal evaluation, the benchmark enables systematic analysis of how models acquire, store, and update real-world knowledge. We release the benchmark, code, and evaluation data to facilitate future research on dynamic factual knowledge in vision–language models.

\section*{Limitations}
The 139 time-sensitive facts obtained from Wikidata refer to frequent entities. Consequently, the results should be interpreted as upper bounds, as factual recall for less common entities may be more difficult and could lead to more outdated or irrelevant results.
Evaluation of editing methods only takes into account the effectiveness of updating time-sensitive facts, but the potential side effects of editing are not evaluated in this work.
Due to limited resources, only open-source models with 7B parameters have been evaluated. As an upper bound on the performance of current VLMs for factual recall, we evaluated closed-source API models.

\bibliography{custom}

\clearpage

\appendix

\section{Appendix}
\label{sec:appendix}

\subsection{Benchmark Construction}\label{app:prompt_implementation_details}
We complement the information in the main paper by providing additional details on the construction of our evaluation benchmark (discussed in Section \ref{sec:v-dyknow}).

\subsubsection{Visual Queries}
For each subject category, we retrieved images from the following Wikidata~\cite{10.1145/2629489} fields: portrait photographs (P18) for athletes; company or institutional logos (P154) for organizations; national flags (P41) and coats of arms (P94) for countries. As prior research indicates that image dimensions can significantly influence model accuracy \cite{rizzoli-etal-2025-civet}, we standardize all images to a fixed resolution of $672 \times 672$ pixels. During this preprocessing phase, uniform black padding was applied where necessary to preserve the target dimensions without distorting the original aspect ratio. Images with transparent backgrounds were rendered with a white background.

Following the image standardization step, we generated prompt templates in which the subject entity is represented by its corresponding image. To construct these queries, we prompted Google Gemini Pro \cite{google_gemini_3-1-pro} to generate five concise question templates for each subject category. Each template is formulated as a question asking for a specific \textit{attribute} (e.g., ``Tim Cook'') of the \textit{subject} entity depicted in the image (e.g., ``Apple''). While the templates differ in their wording, they are intended to express the same underlying query so that they lead to the same attribute value as the answer. To ensure this semantic equivalence, two researchers from our group manually reviewed the generated templates, verifying that each formulation queried the same property and would therefore produce the same attribute as the expected answer. Based on this review, three templates were selected as the final prompts used to query the models. The complete list of prompt templates for our visual queries is reported in Table \ref{tab:visual_prompts}.

\subsubsection{Textual Queries}
For each prompt template used for the visual queries, we derived its corresponding textual counterpart by replacing the part referring the image (e.g., ``this country'', ``this player'', or ``this organization'') with the explicit textual name of the subject entity (e.g., ``this country'' $\rightarrow$ ``Italy'', or ``this organization'' $\rightarrow$ ``Apple''). To guarantee that these modifications did not alter the semantic meaning of the queries, two researchers manually verified the resulting textual prompts, ensuring they remained grammatically correct and accurately reflected the original output response of the visual queries.

\subsubsection{Visual Entity Recognition Prompts}
Following the construction of the visual prompts, we leveraged Google Gemini Pro \cite{google_gemini_3-1-pro} to generate five distinct templates that prompt the models to identify the visual subject entity. Two researchers manually reviewed the generated prompts and selected three for each subject category. The complete set of detection prompt templates is provided in Table \ref{tab:detection_prompts}.

\begin{table*}[h!]
    \centering
\includegraphics[width=\textwidth]{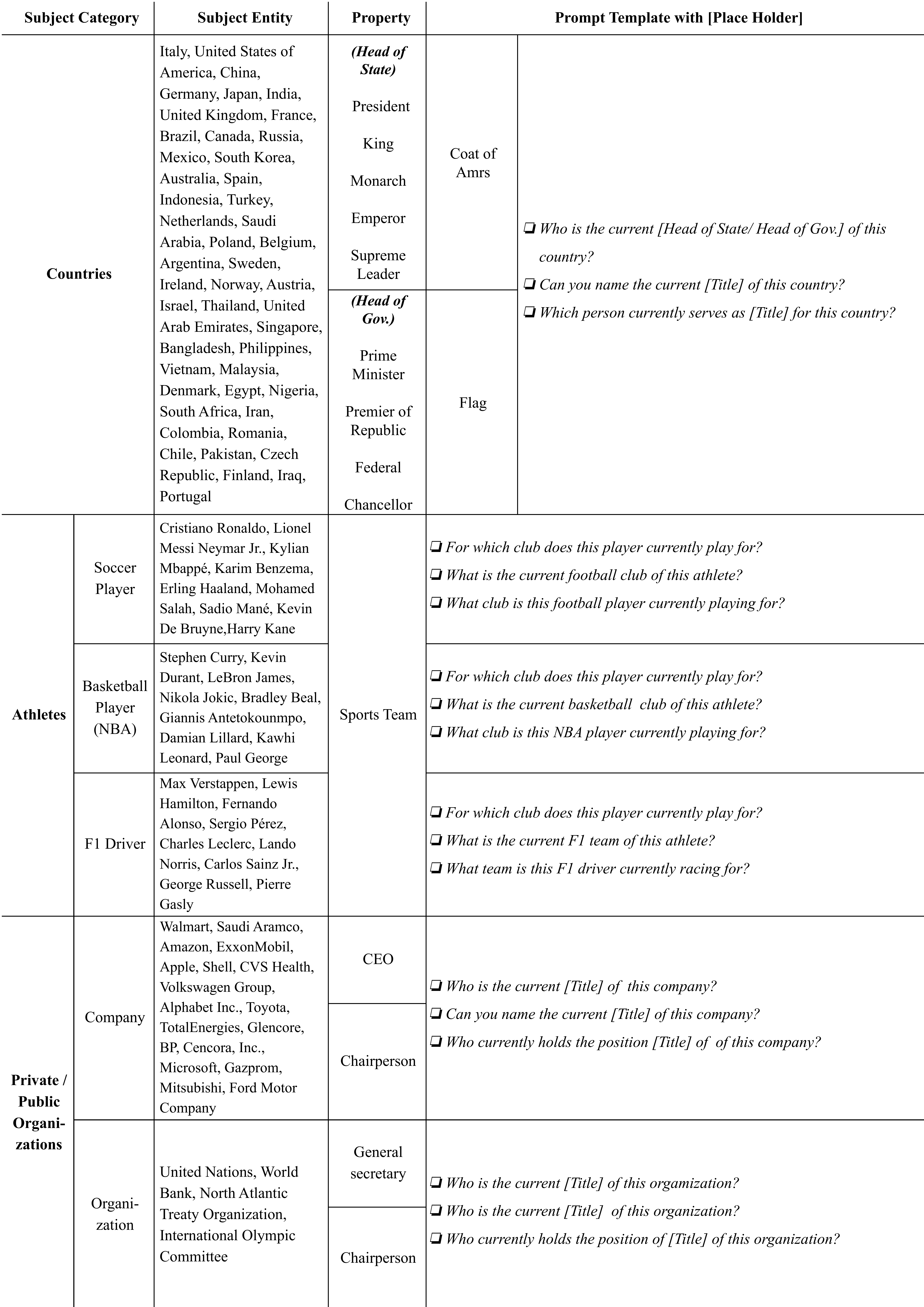}
    \caption{List of all subject entities and their corresponding visual prompts used for benchmarking the VLMs.}
    \label{tab:visual_prompts}
\end{table*}

\begin{table*}[h!]
    \centering
    \includegraphics[width=\textwidth]{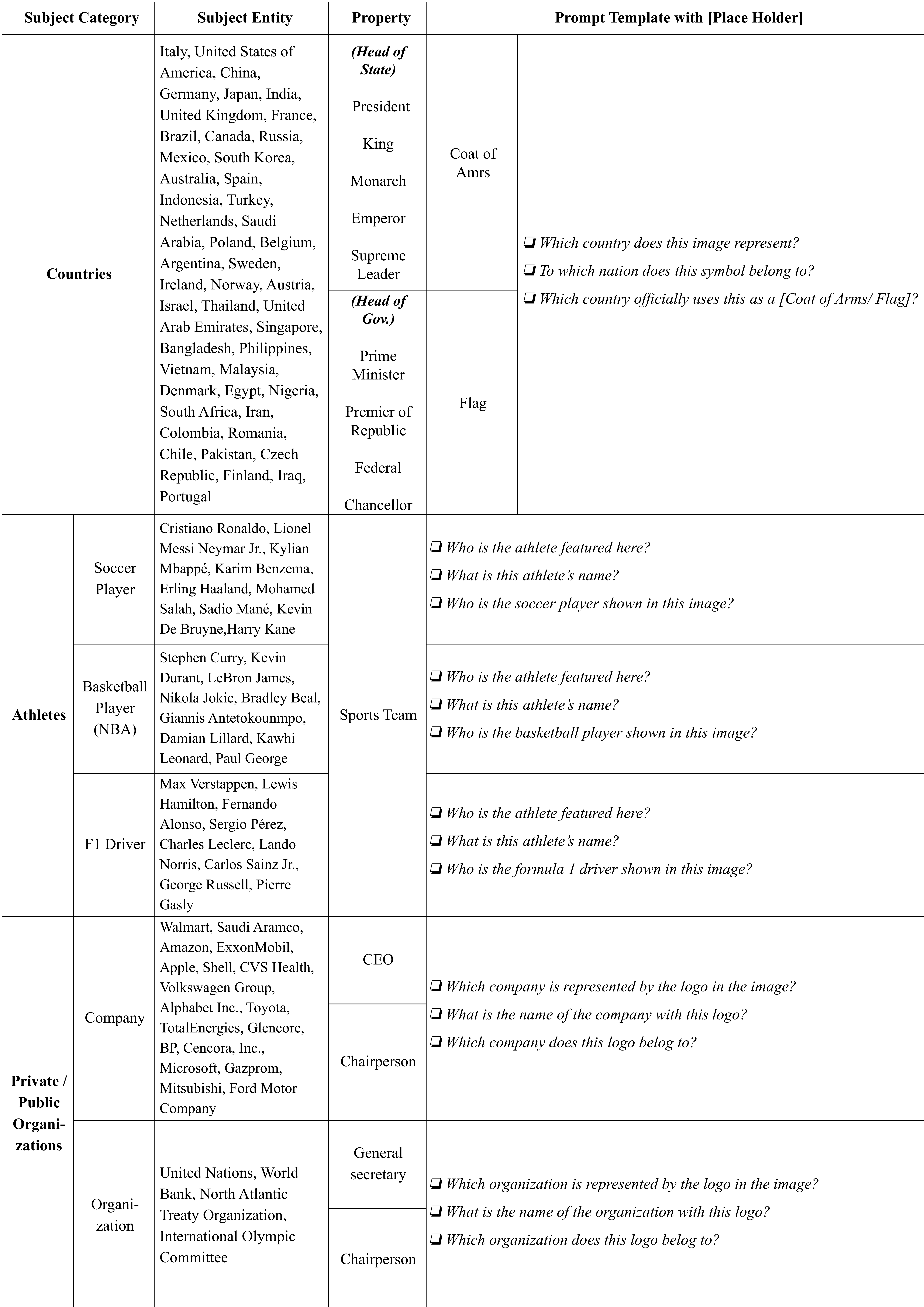}
    \caption{List of all subject entities and their corresponding visual entity recognition prompts used for benchmarking the VLMs.}
    \label{tab:detection_prompts}
\end{table*}

\subsection{Model Checkpoints and Inference Details}\label{app:model_inferece}
To ensure reproducibility, we detail the model checkpoints, API versions, hardware, and generation hyperparameters.

\subsubsection{Model Checkpoints}
We provide the lists of the HuggingFace repository identifiers for the open-weight VLMs and their corresponding LLMs in Table \ref{tab:model_checkpoints}. LLMs baselines were selected to match the base models utilized during VLM training. For closed-source models, we queried the OpenAI using the \texttt{gpt-4.1-2025-04-14} \cite{openai-gpt41-docs} and \texttt{gpt-5.1-2025-11-13} \cite{openai-gpt51-docs} models.

\subsubsection{Inference Hyperparameters}
To ensure deterministic responses, we employed greedy decoding for all open-weight models. To reduce verbose responses, we append the instruction \textit{``Answer with the name only''}, following prior findings~\cite{rizzoli-etal-2025-civet}, and set the maximum number of tokens to 20.

\subsubsection{Hardware}
Open-weight model inference was executed locally on a single NVIDIA RTX 3090 (24GB) GPU using \texttt{bfloat16} precision. Image pre-processing was handled using the default pipeline of each VLM.

\begin{table*}[t]
    \centering
    \small
    \begin{tabular}{lll}
        \toprule
        \textbf{Model Name} & \textbf{VLM Checkpoint} & \textbf{LLM Checkpoint} \\
        \midrule
        LLaVA-1.5 7B & 
        \href{https://huggingface.co/llava-hf/llava-1.5-7b-hf}{\texttt{llava-hf/llava-1.5-7b-hf}} & 
        \href{https://huggingface.co/lmsys/vicuna-7b-v1.5}{\texttt{lmsys/vicuna-7b-v1.5}} \\
        
        LLaVA-OneVision 7B & 
        \href{https://huggingface.co/llava-hf/llava-onevision-qwen2-7b-si-hf}{\texttt{llava-hf/llava-onevision-qwen2-7b-si-hf}} & 
        \href{https://huggingface.co/Qwen/Qwen2-7B-Instruct}{\texttt{Qwen/Qwen2-7B-Instruct}} \\
        
        PaliGemma 2 10B & 
        \href{https://huggingface.co/google/paligemma2-10b-mix-448}{\texttt{google/paligemma2-10b-mix-448}} & 
        \href{https://huggingface.co/google/gemma-2-9b-it}{\texttt{google/gemma-2-9b-it}} \\
        
        Molmo 7B & 
        \href{https://huggingface.co/allenai/Molmo-7B-O-0924}{\texttt{allenai/Molmo-7B-O-0924}} & 
        \href{https://huggingface.co/allenai/OLMo-7B-Instruct-hf}{\texttt{allenai/OLMo-7B-Instruct-hf}} \\
        
        Qwen2-VL 7B & 
        \href{https://huggingface.co/Qwen/Qwen2-VL-7B-Instruct}{\texttt{Qwen/Qwen2-VL-7B-Instruct}} & 
        \href{https://huggingface.co/Qwen/Qwen2-7B-Instruct}{\texttt{Qwen/Qwen2-7B-Instruct}} \\
        
        Qwen2.5-VL 7B & 
        \href{https://huggingface.co/Qwen/Qwen2.5-VL-7B-Instruct}{\texttt{Qwen/Qwen2.5-VL-7B-Instruct}} & 
        \href{https://huggingface.co/Qwen/Qwen2.5-7B-Instruct}{\texttt{Qwen/Qwen2.5-7B-Instruct}} \\
        
        InternVL3.5 8B & 
        \href{https://huggingface.co/OpenGVLab/InternVL3_5-8B-HF}{\texttt{OpenGVLab/InternVL3\_5-8B-HF}} & 
        \href{https://huggingface.co/Qwen/Qwen3-8B}{\texttt{Qwen/Qwen3-8B}} \\
        
        \bottomrule
    \end{tabular}
    \caption{Official HuggingFace model checkpoints for the open-weight VLMs and their corresponding base LLMs used in our evaluation.}
    \label{tab:model_checkpoints}
\end{table*}

\subsection{Editing Methods Implementation Details}\label{app:editing_methods}
We present more details on the editing methods used in the paper:
\begin{enumerate}
    \item \textbf{WISE} \cite{wang2024wise} introduces a dual-memory architecture that partitions new knowledge in a side memory (to store new edits) while maintaining existing pretrained knowledge in the primary memory. At inference time, a router is used to decide which memory to use for an incoming query. If there is no match between the query and the edit scope, the primary memory is selected to generate the final output. In case of a match, the prompt is routed through the side memory to generate the response grounded on the edited knowledge.

    \item \textbf{GRACE} \cite{hartvigsen2023aging} introduces an Adaptor at a chosen model layer to cache targeted edits while leaving the pretrained weights frozen. During the editing phase, the Adaptor saves the hidden activation of the desired editing fact as a key and learns its corrected output as a corresponding value. Each key is assigned a dynamic search radius that expands or splits to prevent overlaps and encourage generalization to similar inputs. At inference time, the Adaptor evaluates the latent distance between the new prompt and the cached keys. If the prompt falls inside a key's radius, the Adaptor intercepts the forward pass and outputs the learned corrected value. If there is no match, the input passes through the original pre-trained weights to generate the final response.
    
    \item \textbf{IKE} \cite{zheng2023can} introduces an in-context learning approach to edit a model's factual knowledge without altering any of its pre-trained weights. To apply an edit, the method retrieves relevant demonstration examples from a training corpus to guide the editing of knowledge. These demonstrations are specifically formatted into three types: i) copy, to introduce the new fact; ii) update, to help the model generalize the injected knowledge; and iii) retain, to guide the model in preserving unrelated facts. These demonstrations, along with the actual up-to-date fact, are prepended directly to the user's target query to condition the model's final answer. However, this technique does not represent a realistic scenario, as it requires the relevant, up-to-date fact to be deterministically provided to the model for each target question.
\end{enumerate}

\begin{table}[t]
    \centering
    \small
    \begin{tabular}{lccc}
        \toprule
        \multirow{2}{*}{\textbf{Model}} & \multicolumn{3}{c}{\textbf{\# Demonstrations (K)}} \\
        \cmidrule(lr){2-4}
        & \textbf{1} & \textbf{2} & \textbf{3} \\
        \midrule
        LLaVA-1.5 & 88.3\perc & 95.6\perc & 0\perc \\
        Qwen2-VL  & 100\perc  & 100\perc  & 99.1\perc \\
        \bottomrule
    \end{tabular}
    \caption{Harmonic mean for IKE with up to $K=3$ demonstrations. Performance peaks at $K=2$, while adding a third demonstration degrades results, suggesting difficulty in retrieving the relevant information from the context. LLaVA-1.5 is an extreme case, as it ends up repeatedly generating \texttt{\textbackslash n} tokens for $K=3$.}
    \label{tab:ike_results}
\end{table}

For WISE and GRACE, we used the implementations provided in the EasyEdit framework \cite{wang2023easyedit}, adopting the default configurations and hyperparameters. As LLaVA-1.5 is not directly supported by EasyEdit, we adapted the configuration designed for LLaVA-OneVision, which belongs to a similar model family. Since both methods require training additional layers and a router module, we performed the training on a single NVIDIA A100 (80\,GiB) GPU.

For IKE, we randomly selected up to $K=3$ demonstrations from the pool of unedited facts. The results in Table~\ref{tab:ike_results} show that, while using two demonstrations improves the harmonic mean, including a third demonstration degrades performance. LLaVA-1.5 represents an extreme case: the model generates only \texttt{\textbackslash n} tokens, possibly because the extended context exceeds its effective context window. Based on these observations, we select the configuration with $K=2$ for the experiments reported in Table~\ref{tab:knowledge_editing_rag_vlm}, as it achieves the highest harmonic mean.

\subsection{Multimodal RAG} \label{mrag}
We provide additional details about the dataset construction and pipeline used to evaluate multimodal RAG.

\subsubsection{Dataset Construction}
We construct a set of image-document pairs for Qwen2-VL and LLaVA-1.5 by collecting passages from Wikipedia. For each outdated entity, we extract passages that contain up-to-date attributes from its Wikipedia page. These passages are manually verified to ensure they contain the correct answer, and one passage is selected. We then pair the selected passage with the main figure from the corresponding Wikipedia page.

\subsubsection{Retrieval and Reranking Pipeline}
We implement the multimodal RAG pipeline using the Qwen3-2B Embedder and Reranker \cite{zhang2025qwen3}. We encode each image-document pair using the Qwen3-2B Embedder and retrieve the 10 candidates with the highest cosine similarity based on the visual query. The retrieved candidates are then re-ranked using the Qwen3-2B Reranker. 

To determine the optimal number of documents $K$ to include in the context, we evaluate configurations with $K \in \{1,3,5\}$. Results in Table~\ref{tab:rag_results} show that increasing the number of documents in the context degrades performance for both models, potentially because additional documents introduce irrelevant or distracting information. For the experiment reported in Table~\ref{tab:knowledge_editing_rag_vlm}, we therefore select the configuration with $K=1$, as it achieves the highest harmonic mean.

\begin{table}[t]
    \centering
    \small
    \begin{tabular}{lccc}
        \toprule
        \multirow{2}{*}{\textbf{Model}} & \multicolumn{3}{c}{\textbf{\# Documents (K)}} \\
        \cmidrule(lr){2-4}
        & \textbf{1} & \textbf{3} & \textbf{5} \\
        \midrule
        LLaVA-1.5 & 73.5\perc & 44.7\perc & 8.7\perc \\
        Qwen2-VL  & 80.1\perc  & 69.8\perc  & 67.7\perc \\
        \bottomrule
    \end{tabular}
    \caption{Harmonic mean for RAG when considering $K \in \{1,3,5\}$ documents as context.}
    \label{tab:rag_results}
\end{table}

\begin{table*}[t]
    \centering
    \includegraphics[width=0.97\textwidth]{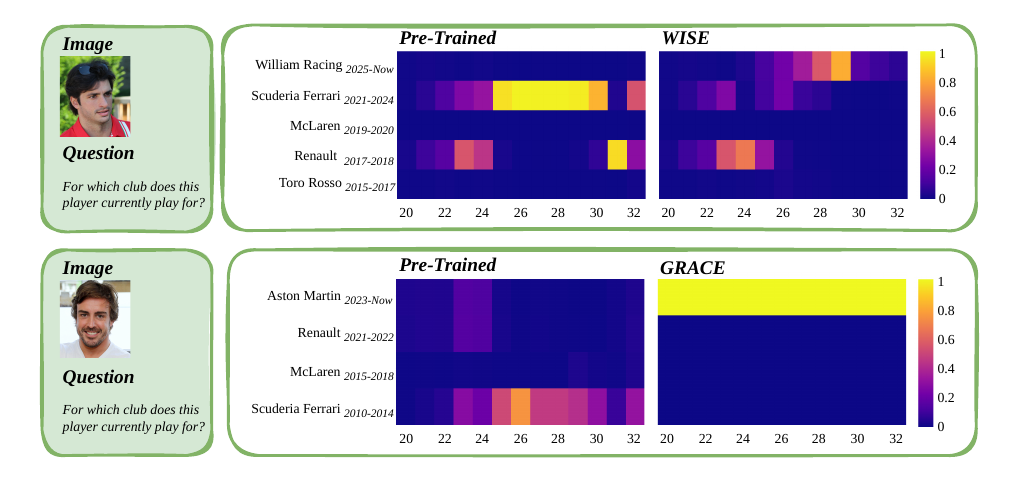}
    \caption{Mechanistic analysis of LLaVA-1.5 illustrating how editing methods \textcolor{green}{successfully} modify the contribution of different layers during factual recall.}
    \label{tab:mechanistic_llava_correct}
\end{table*}

\begin{table*}[t]
    \centering
    \includegraphics[width=0.97\textwidth]{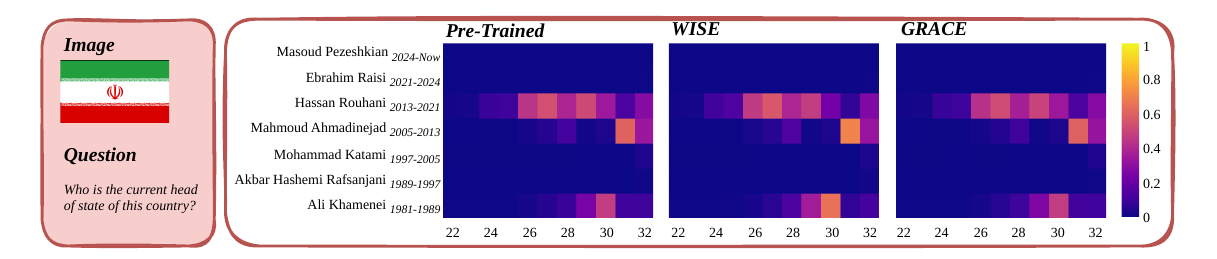}
    \caption{Mechanistic analysis of LLaVA-1.5 illustrating how editing methods \textcolor{red}{fail} to modify the contribution of different layers during factual recall.}
    \label{tab:mechanistic_llava_wrong}
\end{table*}

\subsection{Evaluation Metrics}
\label{subsec:metrics}
Following prior work~\cite{mousavi-etal-2024-dyknow, mousavi2025llms, meng2022locating}, we assess the effectiveness of each approach using the harmonic mean of efficacy and paraphrase success.
Given a time-sensitive question $q$ about a subject-property pair and its up-to-date attribute $a$, efficacy success measures the proportion of cases in which the edited model $\mathcal{M}$ returns $a$ for $q$ (i.e., $\mathcal{M}(q) = a$). Instead, paraphrase success measures the proportion of cases in which $\mathcal{M}$ returns $a$ when prompted using a paraphrased version $q'$ of the same question $q$ (i.e., $\mathcal{M}(q') = a$).

\subsection{Mechanistic Analysis}
\label{subsec:app-mechanistic}
We complement the mechanistic analysis presented in Section~\ref{subsec:mechanistic_interpretability} by reporting the results for LLaVA-1.5. As none of the editing algorithms successfully update the same fact, we select one successful edit for WISE and one for GRACE and report the results in Figure~\ref{tab:mechanistic_llava_correct}. In contrast to Qwen2-VL, a larger number of layers contribute to recalling factual information, suggesting that such knowledge may be stored differently across VLMs. Similar to Figure~\ref{fig:mechanistic}, GRACE propagates the modification across layers 20 to 32; notably, this method also significantly reduces the probability of the other attributes. In contrast, WISE mainly affects layer 29, after which the probability gradually decreases in subsequent layers. Interestingly, the strongest effect does not occur at the editing layer specified in the configuration of WISE, as the method targets layer 23. Moreover, the information associated with the other attributes is not completely removed, suggesting a more localized editing effect.

Figure~\ref{tab:mechanistic_llava_wrong} presents an example of an unsuccessful edit. In this case, both methods show little to no change compared to the activations of the pre-trained model. Interestingly, the probability of the outdated attributes slightly increases for WISE, similarly to the effect described for Qwen2-VL in Figure~\ref{fig:mechanistic}.

\end{document}